\documentclass[11pt,uspaper]{article}

\usepackage{authblk}
\usepackage{setspace}

\usepackage{amsmath}
\usepackage{times}
\usepackage{bm}
\usepackage{natbib}
\usepackage{graphicx}
\usepackage{bbm}
\usepackage[plain,noend]{algorithm2e}

\usepackage{path}
\usepackage{amsmath}
\usepackage{bbm}
\author[1]{P. Dellaportas\thanks{p.dellaportas@ucl.ac.uk}}
\author[2]{A. Plataniotis\thanks{taplat@aueb.gr}}
\author[3]{M. K. Titsias\thanks{mtitsias@aueb.gr}}
\affil[1]{Department of Statistical Science, University College,
 Gower Street, London WC1E 6BT, United Kingdom}
\affil[2]{Department of Statistics, Athens University of Economics and Business,
        Patission 76, Athens 10434, Greece}
\affil[3]{Department of Informatics, Athens University of Economics and Business,
        Patission 76, Athens 10434, Greece}

\begin{document}

\title{\bf Scalable inference for a full multivariate stochastic volatility model}

\maketitle

\begin{abstract}
We introduce a multivariate stochastic volatility model for asset returns that imposes no restrictions to the structure of the volatility matrix and treats all its elements as functions of latent stochastic processes. 
When the number of assets is prohibitively  large, we propose a factor multivariate stochastic volatility model in which the variances and correlations of the factors evolve stochastically over time. Inference is achieved via a carefully designed feasible and scalable Markov chain Monte Carlo algorithm that combines two computationally important ingredients: it utilizes invariant to the prior Metropolis proposal densities for simultaneously updating all latent paths and has quadratic, rather than cubic, computational complexity when evaluating the multivariate normal densities required. We apply our modelling and computational methodology to $571$ stock daily returns of Euro STOXX  index for data over a period of $10$ years.  
MATLAB software for this paper is available at http://www.aueb.gr/users/mtitsias/code/msv.zip.
\end{abstract}


\section{Introduction} \label{intro}

We aim to model a sequence of high dimensional $N \times N$
volatility matrices $\{ (\Sigma_t)_{t=1}^T \}$ of an
N-dimensional zero mean, normally distributed, time series vector of asset returns 
$\{ (r_t)_{t=1}^T \}$. The prediction of $\Sigma_{T+1}$ is a fundamental problem in financial statistics  that has received a lot of attention in portfolio selection and financial management literature, see for example \citet{Tsay05}.  The major statistical challenge emanates from the fact that each $\Sigma_t$ is positive-definite and its number of parameters grows quadratically in $N$.  A popular paradigm in financial econometrics is to adopt observational-driven models that extend the popular univariate GARCH-type formulations, see for example  \citet{Engle02}. In these models 
parameters are deterministic functions of lagged dependent variables so are perfectly predictable one-step-ahead given past information.  We focus, instead, on parameter driven models that assume that  $\{ \Sigma_t \}$ vary over time as dynamic processes with
idiosyncratic innovations. 

The starting point in our model construction is the one-dimensional stochastic volatility model
introduced by  \citet{Taylor86}  which allows the log-volatility of the
observations to be an autoregressive unobserved random process.  
The challenging extension to the multivariate case is discussed in the reviews by  \citet{platanioti2005review}, \citet{Asai06} and \citet{chib2009multivariate}.  
Due to both the computational complexity that increases dramatically with $N$ and the modelling complexity produced by the necessity to stochastically evolve correlations and volatilities preserving the positive definiteness of $\Sigma_t$, all existing models assume some form of model parsimony that often correspond to the simplifications suggested in the observation driven models literature.  In particular, the existing multivariate stochastic volatility (MSV) models assume either constant correlations over time or some form of dynamic correlation modelling through factor models with factors being independent univariate stochastic volatility models;   see, for example, \citet{harvey1994multivariate}, \citet{kim1998stochastic}, \citet{pitt1999time},\citet{bauwens2006multivariate},  \citet{tims2003range}. Different approaches to MSV models have been suggested by \citet{philipov2006factor,philipov2006multivariate} who suggest modelling $\Sigma_t$ as an inverted Wishart process and by \citet{carvalho2007dynamic} who proposed dynamic matrix-variate graphical models. 

We propose a new MSV modelling formulation which is full in the sense that all $N(N+1)/2$ elements of $\Sigma_t$ evolve in time.  
A key idea of our approach is to assume Gaussian latent processes for functions of the eigenvalues and rotation angles of $\Sigma_t$.  By invert-transforming back to $\Sigma_t$ the positive definiteness is immediately ensured. For a $N$-dimensional vector of responses, we construct a MSV model with $N(N+1)/2$ Gaussian latent  paths corresponding to $N$ eigenvalues and $N(N-1)/2$  rotation angles. When $N$ is prohibitively large, we propose a dynamic factor model in which the  volatility matrices of the factors are treated exactly as $\{ \Sigma_t \}$ in the MSV model.  This generalises the existing assumption of factor independence that is prominent in dynamic factor models in many statistical areas including, except of financial econometrics, economics, see for example \citet{forni2000generalized}, and psychology, see for example \citet{ram2013dynamic}.

Although the above model formulation allows the construction of
latent processes ensuring the positive definiteness of $\{ \Sigma_t \}$, the estimation process remains a computationally challenging task.
In practical quantitative finance areas such as portfolio construction and risk management,  interest lies in applications 
where the number of assets $N$ is in the size of hundreds. Our approach is Bayesian so our view to the problem is that we deal with a non-linear likelihood function with a latent $TN(N+1)/2$-dimensional  Gaussian prior distribution. Since the likelihood itself requires evaluation of  a $TN(N+1)/2$-dimensional Gaussian density,  computational efficiency is a major impediment not only because of the cubic computational complexity required to perform the Gaussian density matrix manipulations, but also because Markov chain Monte Carlo (MCMC) algorithms require carefully chosen simultaneous updates of the latent paths so that good chain mixing is achieved.

Our proposed Bayesian inference is carefully designed to handle both these problems. The crucial MCMC moves that update the latent paths are based on an auxiliary Langevin sampler  suggested by \citet{Titsias2011}.    Moreover, we provide algorithms that achieve computational complexity of squared, rather than  cubic, order for the evaluation of the likelihood Gaussian density and its derivative with respect to rotation angles and eigenvalues.  This overcomes a very crucial impediment that is common in many multivariate statistics applications, see for example \citet{banerjee2008gaussian} for a recent review of this problem in spatial statistics. 

We illustrate our method to a computationally challenging, real data example based on ten years daily returns of $571$ stocks of the Euro STOXX  index.  We formulate a factor MSV model and evaluate the predictive ability of a series of models by  gradually increasing the number of factors and evaluating the distance between the predictive volatility matrix and the quadratic covariation of the next day based on $5$-minutes intra-day data.  

\section{The basic multivariate stochastic volatility model}

We assume that the observed asset returns $r_t$ are $N(0,\Sigma_t)$-distributed and that $r_t$ are covariance stationary so $E(\Sigma_t)=\Sigma$ exists.  
The spectral decomposition $\Sigma_t=P_t \Lambda_t P_t^T$ parametrises the $N(N+1)/2$ independent time-changing entries of $\Sigma_t$ to $N$ eigenvalues $\{ ( \Lambda_{it} )_{i=1}^N \}$ 
and $ N(N-1)/2$
parameters in the eigenvector matrices $P_t$. We further write each $P_t$ as  a product of $N(N-1)/2$ Givens
rotation matrices $P_ t = \prod_{i<j} G_{ij}(\omega_{ij,t})$
where the elements of each Givens matrix $G_{ij}(\omega_{ij,t})$ are given by 
\begin{equation}  
G_{ij}[k,l]=\left\{\begin{array}{ll}\phantom{2} \cos{(\omega_{ij,t})}, & \phantom{1} \textrm{if} \phantom{1} k=l=i \phantom{1} \textrm {or}\phantom{1} k=l=j\\
\phantom{2.} \sin(\omega_{ij,t}), & \phantom{1} \textrm{if} \phantom{1} k=i,l=j \\
-\sin(\omega_{ij, t}), & \phantom{1} \textrm{if} \phantom{1} k=j,l=i \\
\phantom{200000}1, & \phantom{1} \textrm{if} \phantom{1} k=l \\
\phantom{200000}0, & \phantom{1} \textrm{otherwise.}
\end{array} \right.  \nonumber
\end{equation}
Each rotation matrix has one parameter, the rotation angle $\omega_{ij,t}$, which appears in only four cells of the matrix.  For each time $t$ there are $N(N-1)/2$ angles 
$\{( \omega_{ij,t}  )_{i<j}\}$ associated with all possible pairs  $(i,j)$ where $i<j,j=1,\ldots,N$.   
We choose $\omega_{ij, t} \in (-\pi/2,\pi/2)$ to ensure uniqueness of the  rotation angles and we transform angles and eigenvalues to $\delta_{ij,t}=\log( \pi/2+\omega_{ij,t}) - \log (\pi/2-\omega_{ij,t})$
and  $h_{i,t} = \log (\Lambda_{it})$.
Our proposed MSV model is 
\begin{eqnarray} \label{MSV}
h_{i,t+1}&=& h_{i,0}+\phi_i^h \cdot (h_{i,t}-h_{i,0})+\sigma^{h}_{i} \cdot \eta_{i,t}^h,~i=1,\ldots,N,~ t=1,\ldots,T-1,  \nonumber \\
\delta_{ij,t+1}&=& \delta_{ij,0}+\phi_{ij}^{\delta} \cdot (\delta_{ij,t}-\delta_{ij,0})+\sigma^{\delta}_{ij} \cdot \eta_{ij,t}^\delta,~i<j,~ t=1,\ldots,T-1, \nonumber \\
h_{i,1} &\sim& N\left( h_{i,0},\frac{(\sigma_{i}^{h})^2}{1-(\phi_i^{h})^2} \right), 
~~~\delta_{ij,1}\sim N
\left(\delta_{ij,0},\frac{(\sigma_{ij}^{\delta})^2}{1-(\phi_{ij}^{\delta})^2}\right), 
\end{eqnarray}
where $|\phi_i^h|<1$ and
$|\phi_{ij}^{\delta}|<1$ are the persistence parameters of each autoregressive process, $\sigma_{i}^h$ and $\sigma_{ij}^{\delta}$ are
corresponding error variances and $\eta_{i,t}^h,\eta_{ij,t}^{\delta}
\sim N (0,1)$ independently. Note  that due to time-changing  prior structure in (\ref{MSV})  our prior is not orthogonally invariant.  The parameter vectors that need to
be estimated are the transformed rotation angles and eigenvalues
$\{ (\delta_t )_{t=1}^{T} \}$,$\{ (h_t)_{t=1}^{T} \}$, and the latent path parameters
$\theta_h =\{(\phi_i^h,h_{i,0},\sigma_{i}^h)_{i=1}^N \}$
 and
 $\theta_\delta = \{(\phi_{ij}^{\delta}, \delta_{ij,0}, \sigma_{ij}^{\delta})_{i<j} \}$
related to transformed eigenvalues and rotation angles respectively.  The volatility matrices $\Sigma_t $ are positive definite since they are obtained by just transforming back the parameters $h_t,\delta_t$ to $P_t$ and $\Lambda_t$.

Givens angles have been used in the past in Bayesian literature in static problems where the focus is improvement of covariance matrix estimation via reference or shrinkage priors; see  \citet{yang1994estimation} and \citet{daniels1999nonconjugate}.  The effect of  left-multiplying a Givens matrix $G_{ij}(\omega_{ij,t})$ to a vector is to rotate the vector clockwise by $\omega_{ij,t}$ radians in the plane spanned by the $ith$ and $jth$ components of the vector.  The covariance matrix $\Sigma_t$ can therefore be viewed as that of a vector of $N$ uncorrelated random variables with variances $(\Lambda_{it})_{i=1}^N$, rotated successively by applying Givens rotation matrices.  Sparsity may be induced by setting many angles equal to zero since when a rotation angle is zero there is no rotation in the corresponding plane.  \citet{cron2016models} exploited this fact and  proposed sparsity modelling of a covariance matrix by placing priors on the Givens angles.

When the assumption of exchangeability between the asset returns is plausible, we suggest using a hierarchical formulation of the form 
\begin{align}
\phi_i^h & = (e^{\tilde{\phi}_i^h} - 1 )/ (e^{\tilde{\phi}_i^h} + 1) \nonumber \\
\tilde{\phi}_i^h | \mu_h, \lambda_h  & \sim \mathcal{N}(\mu_h, \lambda_h^{-1}) \nonumber \\
(\mu_h,\lambda_h)   & \sim \mathcal{N}(\mu_0, (k_0 \lambda_h)^{-1} ) \text{Ga}(\alpha_0, \beta_0). \label{priors}
\end{align}
In the financial applications we are dealing with, this prior specification has great practical importance. In all large portfolios there are assets with fewer observations due to new stock introductions to the market or to an index, mergers and acquisitions, etc. In these cases, the Bayesian hierarchical model allows borrowing strength between persistence parameters which results to their shrinkage towards the overall mean $\mu_h$. Of course, other assumptions such as exchangeability within markets or sectors might be more appropriate and the prior specification may be chosen accordingly. We propose non-informative prior densities for  $\theta_h$ and $\theta_\delta$ by placing an inverse Gamma density for $(\sigma_{i}^h)^2$ and  $(\sigma_{ij}^{\delta})^2$  and an uninformative uniform improper prior density for $h_{i,0}$ and $\delta_{ij,0}$.
Further details, such as the values of the hyperparameters used in our simulations and real data, are given in the Supplementary material. 

\section{The  full factor MSV model}\label{sec:FMSNV}

The basic model (\ref{MSV}) can be extended to a full factor MSV model by assuming that the means of the initial series  $r_t$ are linear combinations of $K$  factors which are modelled as MSV processes.  This can be written as $r_t  =  Bf_t + V^{1/2} \epsilon_t$, and $f_t  \sim  N(0,\Sigma_t)$
where $B$ is a $N \times K$ matrix of factor loadings, $f_t$ is a $K$-dimensional vector that is modelled with the MSV model (\ref{MSV}), $V=  \sigma^2 I$ is an $N \times N$  diagonal matrix of variances and $\epsilon_t$ is a vector of $N$ independent $N(0,1)$ variates.  For identification purposes, constraints on the elements $b_{ij}$ of $B$ must be imposed, so we set $b_{ij}=0$ for $i<j,i \leq K$ and $b_{ii} = 1 $ for $i \leq K$.   The covariance of $r_t$ at time $t$ is separated into systematic and idiosyncratic components $B \Sigma_t B^T + V$. The non-zero values of the factor loadings matrix $B$ are assigned a conjugate Gaussian prior density while the noise variance $\sigma^2$ a standard conjugate inverse Gamma prior; see the Supplementary material for further details. 

The existing factor MSV models assume that $f_t$ are independent univariate stochastic volatility processes, a quite unrealistic assumption given the broad empirical evidence on observed priced factors. We call these models independent factor models.  Our full factor model provides a generalisation by assuming that both  factor variances and correlations evolve stochastically and it reduces to model (\ref{MSV}) when $N=K$, $B=I$ and $\sigma^2=0$ and to an independent factor model by setting all rotation angles equal to zero. 

\section{Estimation}\label{sec:estimation}

To estimate the parameters of the model we follow a fully Bayesian procedure by applying an MCMC algorithm.  We will describe here the algorithmic steps for the full factor MSV model noting that the steps for the simple MSV model are obtained as a special case.  Suppose a set of observed return series vectors $r_t \in \mathbbm{R}^N$ obtained at time instances $t=1,\ldots,T$ that we wish to model by using a full factor MSV model having $K$ latent factors. While in the real application considered in section \ref{sec:appl} we do consider missing values, for notational simplicity next  
we assume that the vectors $r_t$ have no missing values (the treatment of missing values under
the full factor model is straightforward as explained in the previous section). 
The joint probability distribution of all observations, latent variables and parameters is written in the form 
\begin{equation}
\left( \prod_{t=1}^T \mathcal{N}(r_t|B f_t, \sigma^2 I)  \mathcal{N}(f_t|0,\Sigma_t(x_t)) \right) p(X| \theta_h, \theta_{\delta}) p(\theta_h, \theta_{\delta}) p(B,\sigma^2), \nonumber
\end{equation}     
where $x_t = \{(h_{i,t})_{i=1}^p, (\delta_{ij,t})_{i<j}\}$ denotes the $K (K +1)/2$ vector of all transformed angles and log-eigenvalues that determine the volatility matrix at time $t$. The expression $\mathcal{N}(r_t|B f_t, \sigma^2 I)$ represents the density function $\mathcal{N}(B f_t, \sigma^2 I)$ evaluated at $r_t$.  Finally, $X = (x_1,\ldots,x_T)$ denotes the full set of latent variables, represented as a 
row-wise unfolded vector of the $K(K+1)/2 \times T$ matrix in which each $T$-dimensional row vector stores the latent variables associated with a specific Gaussian autoregressive process. Thus, $p(X| \theta_h, \theta_{\delta})$ can be a huge high-dimensional Gaussian distribution, having an inverse covariance matrix with $K(K+1)/2$ separate blocks associated with the independent latent Gaussian processes and where each $T$-dimensional block has a sparse tridiagonal form. 

Performing MCMC for the above model is extremely challenging due the huge state space. For instance, for a typical real world dataset as the one we consider 
in our experimental study, the number of latent variables in $X$ can be of order of millions, for example for $K=50$ and $T=2000$ the size of $X$ is 2,55 millions. 
We develop a well-mixing computationally scalable MCMC procedure 
that uses an effective move that jointly samples (in a single step) all random variables in $X$. 
       
\subsection{The general structure of the  MCMC algorithm}\label{sec:mcmc}

The random variables we need to infer can be naturally divided into three groups: i) the full factor model parameters and latent variables  
$(B,\sigma^2, f_1,\ldots,f_T)$ that appear in the observation likelihoods, ii) the MSV latent variables $X$ that determine the volatility matrices 
and iii) the hyperparameters $(\theta_h, \theta_{\delta})$ that influence the latent Gaussian prior distribution $p(X| \theta_h, \theta_{\delta})$. We construct a Metropolis-within-Gibbs procedure that sequentially samples each of the above three groups of variables conditional on the others.  Schematically, this is described as 
\begin{align}
B,\sigma^2,f_1,\ldots,f_T & \leftarrow  p(B,\sigma^2, (f_t)_{t=1}^T|\text{rest}) 
\propto \left( \prod_{t=1}^T \mathcal{N}(r_t|B f_t, \sigma^2 I) \mathcal{N}(f_t|0,\Sigma_t(x_t)) \right)  p(B,\sigma^2),\nonumber \\ 
X  & \leftarrow p(X | \text{rest}) \propto  \left( \prod_{t=1}^T \mathcal{N}(f_t|0,\Sigma_t(x_t)) \right) p(X| \theta_h, \theta_{\delta}), \nonumber\\
\theta_h, \theta_{\delta} & \leftarrow p(\theta_h, \theta_{\delta}|\text{rest}) \propto  p(X| \theta_h, \theta_{\delta}) p(\theta_h, \theta_{\delta}). \nonumber
\end{align}
The first step of sampling the full factor model parameters is further split into three conditional Gibbs moves for updating  the factor loadings matrix $B$, the variance $\sigma^2$ and the latent factors $f_1,\ldots,f_T$. This involves simulating from standard conjugate conditional distributions the explicit forms of which are given in the Supplementary material. However, the conjugate Gibbs step for sampling the 
latent factors $f_1,\ldots,f_T$ is rather very expensive for our application, as it scales as $O(T  K^3 )$. Therefore we replace this step
with a more scalable Metropolis within Gibbs step that costs $O(T N K)$ as we detail in Section \ref{sec:samplefactors}.
The third step of sampling $\theta_h$ and $\theta_{\delta}$ also involves
standard procedures: Gibbs moves for the parameters $h_{i,0}$, $\delta_{ij,0}$, $(\sigma_i^h)^2$, $(\sigma_{ij}^{\delta})^2$ and Metropolis-with-Gibbs 
for the transformed persistence parameters of the AR processes; full details are given in the Supplementary material. 
The most challenging step in the above MCMC algorithm is the second one where we need to simulate $X$. This
requires simulating from a latent Gaussian variable model where the high-dimensional $X$ follows a Gaussian prior distribution  
$p(X| \theta_h, \theta_{\delta})$ and then generates the latent factors $F = (f_1,\ldots,f_T)$ through a non-Gaussian density 
$
p(F|X) = \prod_{t=1}^T \mathcal{N}(f_t|0,\Sigma_t(x_t)),
\label{eqFlikelihood} \nonumber
$
where $X$ appears non-linearly inside the volatility matrices. We can think of $p(F|X)$ as the likelihood function in this
latent Gaussian variable model where $F$ plays the role of the observed data. To sample $X$ we have implemented an efficient 
algorithm proposed by \citet{Titsias2011} that we describe in Section \ref{sec:auxLangevin} in detail.  

We emphasize that the usual ordering of eigenvalues is not needed during the sampling process since each sampled value of $x_t$ reconstructs invariantly a sample for $\Sigma_t$. Finally, from a practical perspective, the most interesting posterior summary of the MCMC algorithm is the predictive density of $\Sigma_{T+1}$ which is constructed by transforming all the predictive densities of $x_{T+1}$ produced exactly as described in the very first paper on Bayesian estimation for univariate stochastic volatility models by \citet{jacquier1994bayesian}.

\subsection{Auxiliary Langevin sampling for latent Gaussian variables models \label{sec:auxLangevin}}

The algorithm in \citet{Titsias2011} is based on combining the Metropolis-Adjusted Langevin Algorithm (MALA)
with auxiliary variables in order to efficiently deal with a latent Gaussian variable model. 
The use of auxiliary variables allows to construct an iterative Gibbs-like procedure 
which  makes efficient use of gradient information of the intractable likelihood $p(F|X)$ and  is invariant 
under the tractable Gaussian prior $p(X | \theta_h, \theta_{\delta})$. For the 
remaining of this section we shall simplify our notation by dropping reference to the parameters $\theta_h$ and $\theta_{\delta}$ 
which are kept fixed when sampling $X$, so that the Gaussian prior is written as
$
p(X) = \mathcal{N}(X|M, Q^{-1}), \nonumber
$
where $M$ is the mean vector and $Q$ is the inverse covariance matrix. Suppose that we are at the $n$-th iteration of 
the MCMC and the current state of $X$ is $X_n$. We introduce 
auxiliary variables $U$ that live in the same space as $X$ and are sampled from the following Gaussian density conditional on $X_n$:
\begin{equation}
p(U|X_n) = \mathcal{N}(U|X_n +  \frac{\zeta}{2} \nabla \log p(F|X_n), \frac{\zeta}{2} I), \nonumber
\end{equation}
where $\nabla \log p(F|X_n)$ denotes the gradient of the log likelihood evaluated at the current state $X_n$. 
$U$ injects Gaussian noise into the current state $X_n$ and shifts it by $(\zeta/2) \nabla \log p(F|X_n)$, where $\zeta$ is a step size parameter.  Thus, $X_n$ has moved towards the direction 
where the log likelihood takes higher values and $p(U|X_n)$ corresponds to a hypothetical 
MALA proposal distribution associated with a target density that is solely proportional to the likelihood $p(F|X)$. A difference, however, is that
in this distribution the step size or variance is  $\zeta/2$, while in the regular MALA the variance is $\zeta$. This is because 
$U$ aims at playing the role of an intermediate step that feeds information into the construction of the proposal density  for sampling $X_{n+1}$.  The remaining variance $\zeta/2$ is added in a subsequent stage when a proposal is specified in a way that invariance under the Gaussian prior density is achieved.  More precisely, 
if the target was just proportional to the likelihood $p(F|X)$, then we could propose a candidate state $Y$ given $U$ from 
$Y \sim \mathcal{N}(Y|U, \zeta/2)$ and by marginalizing out the auxiliary variable $U$ we would had recovered 
the standard MALA proposal distribution $\mathcal{N}(Y|X_n +  (\zeta/2) \nabla \log p(F|X_n), \zeta)$. However, since our actual target 
is $p(F|X) p(X)$ and $p(X)$ is a tractable Gaussian term, we modify the proposal distribution 
by multiplying it with this Gaussian distribution so that the whole proposal will become invariant under the prior. The proposed $Y$ is sampled from the proposal density
$$
q(Y|U)  = \frac{1}{\mathcal{Z}(U)} \mathcal{N}(Y|U, \frac{\zeta}{2} I) p(Y) 
 = \mathcal{N}(Y |  (I + \frac{\zeta}{2} Q)^{-1} (U  + \frac{\zeta}{2} Q M), \frac{\zeta}{2} (I + \frac{\zeta}{2} Q)^{-1}) \nonumber
$$
where $\mathcal{Z}(U) = \int \mathcal{N}(Y|U, \frac{\zeta}{2} I) p(Y) d Y$.
A proposed $Y$ is accepted or rejected with Metropolis-Hastings acceptance probability $\text{min}(1,r)$ where  
\begin{align}
r & = \frac{p(F|Y) p(U|Y) p(Y)}
{p(F|X_n) p(U|X_n) p(X_n)} 
\frac{q(X_n|U)}{q(Y|U)} =   
\frac{p(F|Y) p(U|Y) p(Y)}
{p(F|X_n) p(U|X_n) p(X_n)}  
\frac{ \mathcal{Z}(U)^{-1} \mathcal{N}(X_n|U, (\zeta/2) I ) p(X_n) }{ \mathcal{Z}(U)^{-1} \mathcal{N}(Y|U, (\zeta/2) I) p(Y) } \nonumber \\ 
& = \frac{p(F|Y) \mathcal{N}(U| Y  +  (\zeta/2) D_y, (\zeta/2) I) }
{p(F|X_n) \mathcal{N}(U|X_n +  (\zeta/2) D_t, (\zeta/2) I) }  
\frac{ \mathcal{N}(X_n|U, \frac{\zeta}{2} I ) }{ \mathcal{N}(Y|U, \frac{\zeta}{2} I ) }  \nonumber \\ 
& = \frac{p(F|Y)}{p(F|X_n)} 
\exp \left\{ - (U - X_n)^T D_t  + (U - Y)^T D_y  - \frac{\zeta}{4} ( || D_y ||^2 - || D_t ||^2 )  \right\} 
\label{eq:ratio}
\end{align}
where $D_t = \nabla \log p(F|X_n)$,  $D_y = \nabla \log p(F|Y)$ and $|| Z ||$ denotes the Euclidean norm of a vector $Z$.  An important observation in the resulting form of (\ref{eq:ratio}) is that the Gaussian prior terms $p(X_n)$ and $p(Y)$ have been 
cancelled out from the acceptance probability, so their evaluation is not required: the resulting $Q(Y|U)$ is invariant under the Gaussian prior. The basic sampling steps are summarised in Algorithm \ref{al1}.
\begin{algorithm}[!h]
\caption{Auxiliary Langevin Sampler algorithm} \label{al1}
\vspace*{-12pt}
\begin{tabbing}
   \enspace (i) $U \sim \mathcal{N}(U|X_n +  (\zeta/2) D_t, (\zeta/2) I)$\\
   \enspace (ii) $Y \sim \mathcal{N}(Y | (I + (\zeta/2) Q)^{-1} (U + (\zeta/2) Q M), (\zeta/2) (I + (\zeta/2) Q)^{-1})$ and  with \\
\qquad     probability $\min(1,r)$, where $r$ is given by  (\ref{eq:ratio}), $X_{n+1}=Y$ or otherwise $X_{n+1}=X_n$. \\
 \end{tabbing}
\end{algorithm}
A simplified version  is obtained when we ignore the gradient from the likelihood $p(F|X)$. Then, the algorithm reduces to an auxiliary random walk Metropolis which 
is implemented exactly as Algorithm \ref{al1} with the only difference that the gradient vectors $D_t$ and $D_y$ are now equal to zero, leading to simplifications of some expressions; for example,  the probability $r$ reduces to the likelihood ratio.  An elegant property of the above auxiliary sampling procedure is that when the Gaussian prior tends to a uniform distribution by letting $Q \rightarrow 0$, it proposes as standard MALA or to standard random walk Metropolis algorithms. This can be seen by observing that the marginal proposal distribution in step (ii) of Algorithm \ref{al1}
reduces to the previous standard schemes where the underlying target distribution will be proportional to 
the likelihood $p(F|X)$. This suggests that in order to set the step size parameter $\zeta$  we can follow the standard practise in adaptive MCMC, so that for the auxiliary Langevin we can tune $\zeta$ to achieve an 
acceptance rate of around $50-60\%$ and for the auxiliary random walk Metropolis an acceptance rate of $20-30\%$. Empirically, we have found 
that these regions are associated with optimal performance; however, there is no so far a theoretical proof. 

Let us now return to  our application. In order to apply the above algorithm to the full factor MSV model where the size of $X$ can be of 
order of millions, we have to make sure that the computational complexity remains linear with respect to the size of $X$. 
This is made possible because the Gaussian prior $\mathcal{N}(X|M, Q^{-1})$ has a sparse tridiagonal inverse covariance matrix $Q$. Thus, given that $Q$ is tridiagonal, the matrix $(2/\zeta) I + Q$ will also be tridiagonal  and similarly the matrix $L$ obtained from the Cholesky decomposition  $L L^T = ( 2/ \zeta) I + Q$, will be a 
lower two-diagonal matrix 
which can be computed efficiently in linear time. Then, a sample $Y$ in the step 2 
of Algorithm \ref{al1} can be simulated according to  
\begin{equation}
Y =  L^{-T} ( L^{-1} ( \frac{2}{\zeta} U +   Q M )  + Z), \ \ Z \sim \mathcal{N}(0, I), \nonumber
\end{equation}      
where parentheses indicate the order in which the computations should be performed. All these computations, including the two linear 
systems needed to be solved, can be performed efficiently in linear time since the associated matrices are either tridiagonal or 
lower two-diagonal. Therefore, the overall complexity when sampling $Y$ is linear with respect to the size of this vector. Since 
this vector has size $K (K+1)/2 \times T$ the computational complexity scales as $O(T K^2 )$.

Finally, the above algorithm requires the evaluation of the acceptance probability which is dominated by the likelihood ratio 
that involves the density $p(F|X)$ given by  (\ref{eqFlikelihood}) which consists of a product of 
$T$ $K$-dimensional multivariate Gaussian densities. Furthermore, we need to compute gradients of the form $\nabla \log p(F|X)$ of this log likelihood
that appear in the acceptance probability and are required also when sampling $U$.  
A usual computation of these quantities scales as $O(T K^3)$ which is too expensive for the real applications of the full factor MSV model. 
By taking advantage of the analytic properties of the Givens matrices we can reduce the computational complexity to $O(T K^2)$, that is 
quadratic with respect to dimensionality of the Gaussians. To achieve such a complexity we have developed the specialized algorithms  
detailed in the next Section.    

The proposal density for $X$ in the above MCMC algorithm can be viewed as a generalisation of the preconditioned Crank-Nicolson proposals, see \citet{cotter2013mcmc}, which also are designed to be invariant to the prior.  By integrating out the latent variable $U$ one observes that the proposal density in Algorithm \ref{al1} is just a vector autoregressive update of $X_n$ with a vector of coefficients being a complicated function of $D_t$ and $D_y$.  Such a generalisation was suggested by  \citet{law2014proposals}.  What is important in the algorithm by \cite{Titsias2011}, and of primary importance in cases where the dimension of $X$ is large, is that the computational complexity required to apply such a Metropolis proposal is dramatically alleviated through the inclusion of the latent variable $U$.

\section{Computational complexity}

\subsection{$O(K^2)$ computation for the MSV model \label{sec:OK2}}  

A crucial property of the MSV model is that the evaluation of its log density and the corresponding gradients 
with respect to the parameters inside the volatility matrix $\Sigma_t$ can be computed in  
$O(K^2)$ time. This differs with other more commonly-used parametrizations of the
multivariate Gaussian distribution where computations scale as $O(K^3)$ and they are infeasible  
for large $K$.  Assume we wish to evaluate the log density associated with the vector  
$r_t  \sim N(0,\Sigma_t)$ written as 
\begin{equation}
\log \mathcal{N}(r_t|0,\Sigma_t) = - \frac{K}{2} \log (2 \pi) 
- \frac{1}{2} \sum_{i=1}^K h_{it} - \frac{1}{2} v_t^T  v_t,  \label{logN}
\end{equation} 
where $v_t = \Lambda_t^{-\frac{1}{2}} P_t^T r_t$ and where we used that $\log |\Sigma_t| = \log |\Lambda_t| 
= \sum_{i=1}^N h_{it}$. Clearly, given $v_t$ the above expression takes $O(K)$ time to compute. Therefore,    
in order to prove $O(K^2)$ complexity we need to show that the computation of $v_t$ scales as $O(K^2)$. 
This is based on the fact that the transformed vector $G_{ij}(\omega_{ji,t})^T v_t$ takes $O(1)$ time 
to compute since all of its elements are equal to the corresponding ones from the vector $v_t$  
apart from the $i$-th and $j$-th elements that become
 $v_{t}[i] \cos(\omega_{ji,t})  - v_{t}[i] \sin(\omega_{ji,t})$ and  
 $v_{t}[j] \sin(\omega_{ji,t}) + v_{t}[j] \cos(\omega_{ji,t})$, respectively. 
Thus, the whole product with all $K(K-1)/2$ Givens matrices can be 
carried out recursively in $O(K^2)$ time as shown in Algorithm \ref{al2}.
\begin{algorithm}[!h]
   \caption{Recursive algorithm for computing $v_t$ in $O(K^2)$ time.  $\text{diag}(A)$ is the vector of the diagonal elements of a square matrix $A$.}  \label{al2}   
\begin{tabbing}
   \enspace Initialize $v_t = r_t$. \\
   \enspace For $i=1$ to $K-1$ \\
   \quad For $j=i+1$ to $K$    \\ 
    \quad \quad Set $c = \cos(\omega_{ij,t})$, \ $s = \sin(\omega_{ij,t})$ \\
    \quad \quad Set $t_1 = v_{t}[i]$, \  $t_2 = v_{t}[j]$\\
    \quad \quad Set $v_{t}[i] \leftarrow c*t_1  - s*t_2$   \\
    \quad \quad Set $v_{t}[j] \leftarrow s*t_1 + c*t_2$ \\
   \quad End For \\
   \enspace End For \\
   \enspace $v_t = v_t \circ \text{diag}(\Lambda_t^{- 1/2})$ 
\end{tabbing}
\end{algorithm}

The derivatives of the log density (\ref{logN}) with respect to the vector of log eigenvalues $h_t$ is simply 
$ - 1/2  + (1/2) v_t \circ v_t $, where the symbol $\circ$ denotes element-wise product,
and it is computed in $O(K)$ time given that we have pre-computed $v_t$. 
The partial derivative with respect to each rotation angle 
$\omega_{ij, t}$ takes the form 
\begin{equation}
- v_t^T \frac{\partial v_t}{\partial \omega_{ij, t}} = - v_t^T
\Lambda_t^{-\frac{1}{2}} \left( G_{N N-1}^T  \ldots G_{i j-1}^T \right) \frac{\partial G_{ij, t}^T }{\partial \omega_{ij,t} } 
\left( G_{i j+1}^T  \ldots G_{1 2}^T \right) r_t 
= - \alpha_{ij, t}^T \beta_{ij, t} \nonumber
\end{equation}
where $\alpha_{ij, t} = v_t^T \Lambda_t^{-1/2} ( G_{N N-1}^T  \ldots G_{i j-1}^T ) $  
and $\beta_{ij, t} = (\partial G_{ij, t}^T  / \partial \omega_{ij,t} ) ( G_{i j+1}^T  \ldots G_{1 2}^T ) r_t $ and the partial derivative matrix
$(\partial G_{ij, t} / \partial \omega_{ij,t} )$ is very sparse, having only four non-zero elements, given by  
\begin{equation} 
\label{eq:partialGivens}
\frac{\partial G_{ij, t}}{\partial \omega_{ij,t} } [k, l]
=\left\{\begin{array}{ll}\phantom{2} - \sin{(\omega_{ij, t})}, & \phantom{1} \textrm{if} \phantom{1} k=l=i \phantom{1} \textrm {or}\phantom{1} k=l=j\\
\phantom{2.} \cos(\omega_{ij, t}), & \phantom{1} \textrm{if} \phantom{1} k=j, l=i \\
-\cos(\omega_{ij, t}), & \phantom{1} \textrm{if} \phantom{1} k=i, l=j \\
\phantom{200000}0, & \phantom{1} \textrm{otherwise}
\end{array} \right.
\end{equation}
where  $i<j$. All $\alpha_{ij,t}$ and $\beta_{ij, t}$, for $i < j$,  can be computed in 
$O(K^2)$ time by carrying out two separate forward and backward recursions constructed similarly to the Algorithm \ref{al2}. 
Then, all final $K (K-1)/2$ dot products $\alpha_{ij, t}^T \beta_{ij, t}$ that give the derivatives for all Givens 
angle parameters can be computed in overall $O(K^2)$ time by using the fact that $\beta_{ij, t}$ contains only two non-zero elements 
so that an individual dot product $\alpha_{ij, t}^T \beta_{ij, t}$ takes $O(1)$ time.  This is 
due to the fact that the final multiplication in the computation of  $\beta$ is performed with 
the sparse matrix $\partial G_{ij, t} / \partial \omega_{ij,t} $ that has only four non-zero elements.  
A complete pseudo-code of the above procedure is given in the Supplementary material.

\subsection{Sampling the factors in $O(T N K)$ time \label{sec:samplefactors}}

The exact Gibbs step for sampling each latent factor vector $f_t$ scales as $O(K^3)$ 
while sampling all of such vectors requires $O(T K^3)$ time, a cost that is prohibitive for 
large scale multivariate volatility datasets. To see this, notice that the posterior conditional distribution over $f_t$ is written in the form
\begin{equation}
p(f_t|\text{rest}) \propto \mathcal{N}(r_t|B f_t, \sigma^2 I) \mathcal{N}(f_t|0,\Sigma_t), 
\label{eq:condtionalFt}
\end{equation}
which gives the Gaussian $p(f_t|\text{rest}) = \mathcal{N}(f_t | \sigma^{-2} M_t^{-1} B^T r_t, M_t^{-1})$ where $M_t = \sigma^{-2} B^T B + \Sigma_t$. To simulate from this Gaussian we need first to compute the stochastic volatility matrix $\Sigma_t$ 
and subsequently the Cholesky decomposition of $M_t$. Both operations have a cost
$O(K^3)$ since the matrix product $B^T B$, that scales as $O(N K^2)$, needs to be computed once across all time instances and therefore 
will not dominate the computational cost since typically $N \ll T K$. 
Furthermore, given that there is a separate matrix $M_t$ for each time instance we  need
in total $T$ computations of the volatility and Cholesky matrices in each iteration 
of the sampling algorithm, which adds a cost that scales as $O(T K^3)$. 
The matrix-vector products $B f_t$, needed to compute the means of the Gaussians, 
scale overall as $O(T N K)$, but in practice this will be much less expensive than the term $O(T K^3)$. We note here that a matrix multiplication is the simplest computation with little overhead that can be trivially parallelised in modern hardware.
To avoid the $O(T K^3)$ computational cost we replace the exact Gibbs step with a much faster Metropolis within Gibbs step that 
scales as $O(T (N K + K^2))$.  Specifically, given that eq.\ (\ref{eq:condtionalFt}) is of the form of a latent Gaussian model, where
$\mathcal{N}(f_t|0,\Sigma_t)$ is the Gaussian prior and  $\mathcal{N}(r_t|B f_t, \sigma^2 I)$ the (Gaussian) 
likelihood, we can apply the auxiliary Langevin scheme as described in Section \ref{sec:auxLangevin}. By introducing 
the auxiliary random variable $U_t$ drawn from 
\begin{equation}
p(U_t|f_t) = \mathcal{N}(U_t|f_t +  \frac{\zeta_t}{2} D_{f_t}, \frac{\zeta_t}{2} I), \nonumber
\end{equation}
where $D_{f_t} = \nabla \log \mathcal{N}(r_t|B f_t, \sigma^2 I) =  \sigma^{-2} B^T (r_t - B f_t)$, the auxiliary Langevin 
method is applied as shown in Algorithm \ref{al3}.
\begin{algorithm}[!h]
\caption{Auxiliary Langevin for the latent factors} \label{al3}
\vspace*{-12pt}
\begin{tabbing}
   \enspace (i) $U_t \sim \mathcal{N}(U_t|f_t +  (\zeta_t/2) D_{f_t}, (\zeta_t/2) I)$\\
   \enspace (ii) Propose $y \sim \mathcal{N}(y | (2/\zeta_t I + \Sigma_t^{-1})^{-1} (2/\zeta_t) U_t, ((2/\zeta_t) I + \Sigma_t^{-1})^{-1})$ and  accept it with \\
\qquad     probability \\
\qquad \enspace 
$r = \frac{\mathcal{N}(r_t|B y, \sigma^2 I)}{ \mathcal{N}(r_t|B f_t, \sigma^2 I)}  
\exp \left\{ - (U_t - f_{t})^T D_{f_t}  + (U_t - y)^T D_y  - \frac{\zeta_t}{4} ( || D_y ||^2 - || D_{f_t} ||^2 )  \right\}$.  
\end{tabbing}
\end{algorithm}
Now observe that the step for sampling $y$ takes $O(K^2)$ time because the eigenvalue decomposition of the covariance 
matrix $((2/\zeta_t) I + \Sigma_t^{-1})^{-1}$ can be expressed analytically as $P_t ( (2/\zeta_t) I + \Lambda_t^{-1})^{-1} P_t^T$
where $\Sigma_t = P_t \Lambda_t P_t^T$ is  the spectral decomposition of $\Sigma_t$. 
Therefore, we can essentially apply Algorithm \ref{al2} to sample $y$ in $O(K^2)$ time. Furthermore, the most expensive 
operation in the M-H ratio above is the computation of the matrix-vector product $B f_t$ which costs $O(N K)$. 
Therefore, overall for all factors across time we need $O(T (N K +  K^2))$ operations which is typically dominated 
by $O(T N K)$ since $K \ll N$. Thus, by taking advantage of the analytic form of the 
eigenvectors of $\Sigma_t$ the computational complexity for proposing $y$ (when sampling $f_t$) is 
reduced from $O(K^3)$ to $O(K^2)$.  The crucial sharing of eigenvectors property is due to the 
spherical or isotropic covariance matrix $(2/\zeta) I$ that is added to $\Sigma_t^{-1}$ and it does not alter 
the eigenvectors. In contrast, this does not hold for the Gaussian proposal used in Gibbs sampling because 
the inverse covariance matrix of that proposal, given by $M_t = \sigma^{-2} B^T B + \Sigma_t^{-1}$, is obtained
 by adding an non-isotropic covariance matrix $\sigma^{-2} B^T B$ to the matrix $\Sigma_t^{-1}$ which results the 
 eigenvectors of $M_t$ to be different that those of $\Sigma_t$. 

\subsection{Full factor MSV model against an MSV model}

From a computational perspective, the full
factor model is more advantageous than model (\ref{MSV}) because it deals with missing values more efficiently.  Suppose that $N=K$ and that we use the multivariate model in (\ref{MSV}) and the returns vector at time $t$ contains missing values, so that  $r_t = (r_{t,o}, r_{t,m})$ where $r_{t,o}$ is the sub-vector of observed components and $r_{t,m}$ are the missing components. The standard Bayesian treatment is  to marginalize out the unobserved values $r_{t,m}$ and obtain the likelihood term (at time $t$) given by $r_{t,o} \sim N(0,\Sigma_{t,o})$ where $\Sigma_{t,o}$ is the sub-block of the full covariance matrix $\Sigma_t$ that corresponds to the observed dimensions. The computation of $\Sigma_{t,o}$ is expensive since it requires first the computation and storage of the full matrix $\Sigma_t$ which scales as $O(N^3)$. 
Given that we have $T$ such matrices the whole computation scales as $O(T N^3)$ which is prohibitively expensive, especially 
within an MCMC algorithm where these computations are repeated in each iteration. 
A Gibbs step for sampling the missing $r_{t,m}$, instead of marginalizing them out, also suffers from the same computational cost. In contrast, for the full factor model the treatment of missing values is very simple since  $r_{t,o}$ and $r_{t,m}$ are conditionally independent given the latent factors $f_t$ and thus the marginalization of $r_{t,m}$ is trivial.  The computational burden is moved to the MCMC sampling of the latent vectors $f_t$, but as we showed in Section \ref{sec:samplefactors}, sampling of $f_t$ can be achieved in $O(T N K)$ time. 
  
Since in nearly all financial applications of daily asset returns there are many missing values, encountered for example because in multinational market portfolios  holidays differ between countries, we recommend using the full factor model with $N=K$, $B=I$ and $\sigma^2>0$ even when the number of assets $N$ is manageable.

\section{Application \label{sec:appl}}

\subsection{Data and MCMC specifics}

We illustrate our methodology by modelling daily returns from $600$ stocks of the  STOXX Europe 600 Index downloaded from Bloomberg between 10/1/2007 to 5/11/2014. The cleaning of the data involved removing $29$ stocks by requiring for each stock at least $1000$ traded days and no more than $10$ consecutive days with unchanged price.  The final dataset had $N=571$ and $T=2017$.  There were $36340$ missing values in the data due to non-traded days, asynchronous national holidays, etc.  A smaller dataset was created by selecting $1000$ days and the $29$ stocks from the Italian stock market.  This dataset was used for a large empirical study to compare our models with independent factor MSV models.  

All MCMC algorithms in this Section had an adaptive time for burn-in consisted of $10^4$ iterations with resulting acceptance probability of $50-60\%$ for all auxiliary Langevin steps. After this adaptive burn-in phase all proposal distributions are kept fixed and then we further performed $10^4$ iterations to finally collect $10^3$ or $2 \times 10^3$ (thinned) samples for the large and small datasets respectively. 

\subsection{Predictive ability}

We compare the predictive ability of our full factor MSV model against an independent factor MSV model in which all factors are assumed to be uncorrelated.  The basis of our comparison is the  predictive likelihood function.  Each model produces a predictive distribution and therefore a predictive likelihood can be evaluated for a future observation.  The comparison of these predictive likelihoods decomposes the Bayes factor one observation at a time and a cumulative predictive Bayes factor through a future time period serves as a Bayesian evidence in favour of a model based on predictive performance; see, for example, \citet{geweke2010comparing}.  Denote by $X_t$ the dynamic latent path and by $\theta$ all the static parameters of the model. Assume that at time $T$ we have obtained MCMC samples $X^s_{1:T}$ and $\theta^s$, $s=1,2,\ldots,S$ based on observed data $r_{1:T} = (r_1,r_2,\ldots,r_{T})$.  For each model, the one-step-ahead predictive likelihood conditional on $\theta$  is given by
\begin{equation}
p(r_{T+1}|r_{1:T},\theta) = \int p(r_{T+1}|r_{1:T},\theta) d F(X_{T+1} \mid r_{1:T},\theta)  \nonumber
\end{equation}
and a Monte Carlo estimate is obtained as
\begin{equation}
{\hat p(r_{T+1}|r_{1:T},\theta)} = S^{-1} \sum_{s=1}^S p(r_{T+1}|X^s_{T+1},\theta)   \nonumber
\end{equation}
where $X^s_{T+1}$ are samples from the transition densities (\ref{MSV}).  This procedure is repeated producing $M$ one-step-ahead predictive likelihoods for data $r_{T+1:T+M}$ where $\theta$ is kept fixed at the sample mean $S^{-1} \sum_{i=1}^{S} \theta^s$ and the required 
samples from the density  $f(X_{t} \mid r_{1:t}, X_{1:t-1})$ for $t=T+1,\ldots,T+M-1$ are obtained through the auxiliary particle filter of \citet{pitt1999filtering}.  Thus, marginal likelihoods for each model can be obtained through
\begin{equation}
{\hat p(r_{1:T+M}|r_{1:T},\theta)} = \prod_{t=T+1}^{T+M} {\hat p(r_{t+1}|r_{1:t},\theta)}   \nonumber
\end{equation}
and the predictive Bayes factors in favor of one model against the other can be readily calculated, see \citet{geweke2010comparing}, \citet{pitt1999time}.

\begin{figure}[!htb]
\centering
{\includegraphics[width=\textwidth]{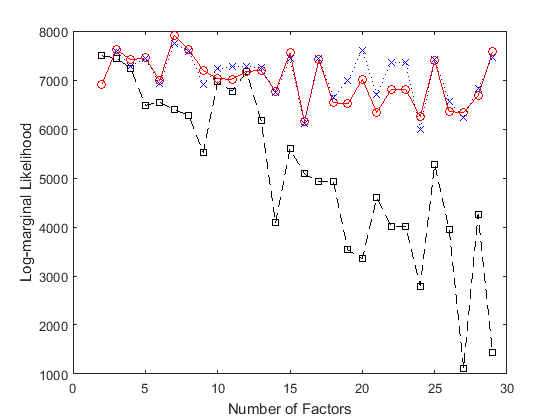}}
\caption{Logarithm of marginal likelihood.  Solid, circle: full factor model with exchangeable priors; Dotted, cross: full factor model with independent priors;Dashed, square: Independent factor model.  }
\label{logm}
\end{figure}

The smaller dataset based on $29$ stocks was selected by excluding the last $100$ days of the larger dataset such that $T=1000$ and $M=100$,  representing a predictive period of about $4$ months.  Marginal likelihoods were calculated for all full and independent factor MSV models.  For the class of independent factor MSV models the best model turned out to be the 2-factor model with estimated log-Bayes factor 50.9 against the second best model with $3$ factors. Recall that a log-Bayes factor greater than $5$ is considered to be a very strong evidence in favour of one model against the other, see \citet{kass1995bayes}.  Actually the marginal likelihood decreased with the number of factors, implying that the dynamic nature of the covariance structure cannot be captured by increasing the size of independent latent processes, see Figure \ref{logm}.  This also reflects the inherent limitation of the independent factor models that attempt to estimate dynamic correlations through (static) linear combinations of univariate independent stochastic processes.   More latent processes just add further noise resulting to decrease of Bayes factors.

The best full factor MSV model turned out to be the $7$-factor model with log-Bayes factor $385.24$ against the independent $2$-factor MSV model.  The largest $29$-factor model had a log-Bayes factor of $71.10$ in favour against the best independent factor model.    There is overwhelming evidence that full factor models provide better predictions.  Moreover, the general picture of Figure \ref{logm} indicates that the full factor MSV model is more robust, across number of factors, when compared with the independent factor model.  

Figure \ref{logm} also depicts the log-marginal likelihood of the full factor model in which independent $N(0,10^3)$ priors were used for $\tilde{\phi}_i^h | \mu_h, \lambda_h $ instead of the exchangeable priors in (\ref{priors}).  The model with $7$ factors achieved again the highest marginal likelihood  but with lower value than that obtained by the exchangeable prior by 155.58, indicating that for this data the exchangeable prior produces a significantly preferable model with respect to the predictive Bayes factor. 

\subsection{Computational efficiency}

\begin{figure}[!htb]
\centering
{\includegraphics[width=\textwidth]{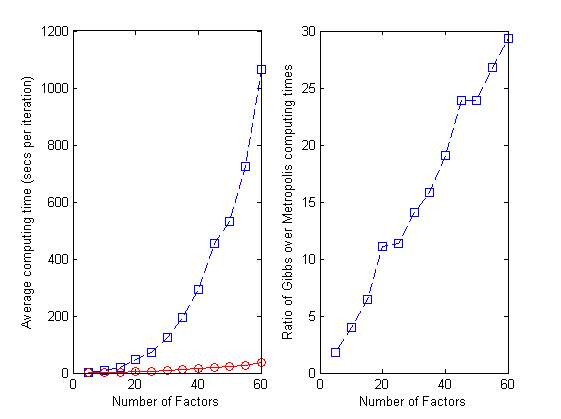}}
\caption{Left panel: Average computing time for Metropolis-Hastings (circle) and Gibbs sampling (square) algorithms.  Right panel: Computing time ratio between Gibbs sampling and Metropolis-Hastings algorithms.}

\label{times}
\end{figure}

One may question whether the increased efficiency achieved by sampling the factors with the Metropolis sampler of Section \ref{sec:samplefactors} achieves a realistically faster algorithm than the simple Gibbs sampling algorithm.  Figure \ref{times} presents how computation time scales with number of factors in the large dataset.  The left panel illustrates that for large problems the Gibbs sample has a prohibitive computational cost, whereas the right panel demonstrates how the computing time ratio increases with the number of factors.  

In smaller examples with few factors in which the computation of the Gibbs sampler is feasible, it is of interest to inspect the Markov chain mixing of the two algorithms. For the small dataset, Table \ref{t1}  presents the effective sample sizes and the computing times for the $7$-factor full MSV model with the Metropolis and Gibbs samplers. The parameters inspected are the $29 \times 30 / 2$ elements of the $29$-dimensional covariance matrix $\Sigma_T$ based on  $10,000$ unthinned iterations.  The computing times are comparable, but the Metropolis algorithm clearly outperforms the Gibbs sampler in terms of Markov chain convergence efficiency.

\begin{table}[!htb]
\centering
		\begin{tabular}{ccccc}
			\\
			 Method & Time(s) & Minimum ESS & Maximum ESS & s / Minimum ESS \\[5pt]
 Metropolis & 2954.2 & 3025.3 & 6538.7 & 0.98 \\
 Gibbs & 3339.2 & 122.2 & 4164.3 & 27.33
		\end{tabular}
		\label{t1}
		\caption{Effective sample sizes (ESS) and computing times in seconds (s) for sampling the factors in the 7-factor full MSV model}
			\end{table}

\subsection{Application to a large dataset}

To apply our full factor MSV model to the large dataset we need to choose the number of factors $K$.  The scale of the problem makes the calculation of  marginal likelihoods for each $K$ computationally infeasible, so we propose comparing one-step ahead forecasts of different $K$ against a proxy. We use as a proxy for $\Sigma_{T+1}$  the realized covariation matrix calculated as the cumulative cross-products of five minutes intraday returns; see
\citet{Andersen99} and ~\citet{Barndorff-Nielsen04}.
If an element of an $N \times N$ covariance matrix $\sigma_{ij}$ is estimated by the elements of the posterior mean of $\Sigma_{T+1}$ with elements  ${\hat \sigma}_{ij}$
and its corresponding proxy estimate is $\sigma_{ij}^*$, we use as discrepancy measures to test how competing models perform the mean absolute deviation given as 
$N^{-2} \sum_{i,j} |\sigma^{*}_{ij}-\sigma_{ij}| $ and the root mean square error given by $ [ N^{-2} \sum_{i,j}  (\sigma^{*}_{ij}-\sigma_{ij})^2 ]^{1/2} $.  

For $K=20,30$ and $40$ the corresponding values of these quantities were $(0.0605,0.0601,0.0635)$ and $(0.0567,0.0553,0.0614)$ respectively, so there is an indication that out of sample forecasting ability of the STOXX 600 volatility matrix is better with around $K=30$  factors.  Sometimes prediction of more days ahead might be of interest, for example when portfolio re-allocation is performed in different time scales, so we also predicted $\Sigma_{T+2}$ and again the corresponding discrepancy measures were $(0.0763,0.0720,0.0811)$ and $(0.0775,0.0697,0.0867)$ respectively, verifying that $K=30$ factors have a comparatively better predictive ability.  

\begin{figure}[!htb]
\centering
{\includegraphics[width=\textwidth]
{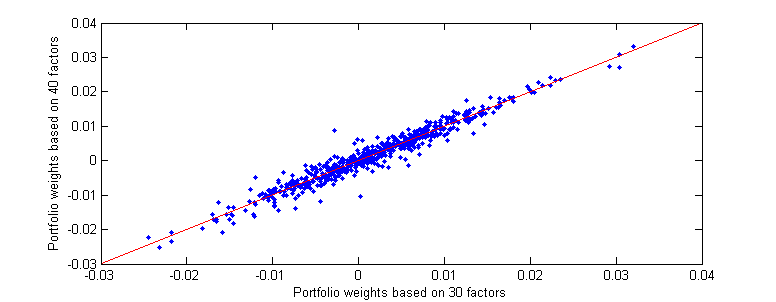}}
\caption{Next day minimum variance portfolio weights of 571 stocks of {\em STOXX Europe 600 index}   based on 30 factors against those based on 40 factors.}
\label{weights}
\end{figure}

Figure \ref{weights} presents the $571$ minimum variance portfolio weights with $30$ and $40$ factors calculated as $\Sigma_{T+1}^{-1} \iota / \iota{'} \Sigma_{T+1}^{-1} \iota$ where $\Sigma_{T+1}$ is estimated with the MCMC-based posterior predictive mean and $\iota$ is an $N \times 1$ vector of ones.  It is clear that the magnitude of the weights remains considerably constant especially in the financially important values away from zero. 

\begin{figure}[!htb]
\centering
{\includegraphics[width=\textwidth]
{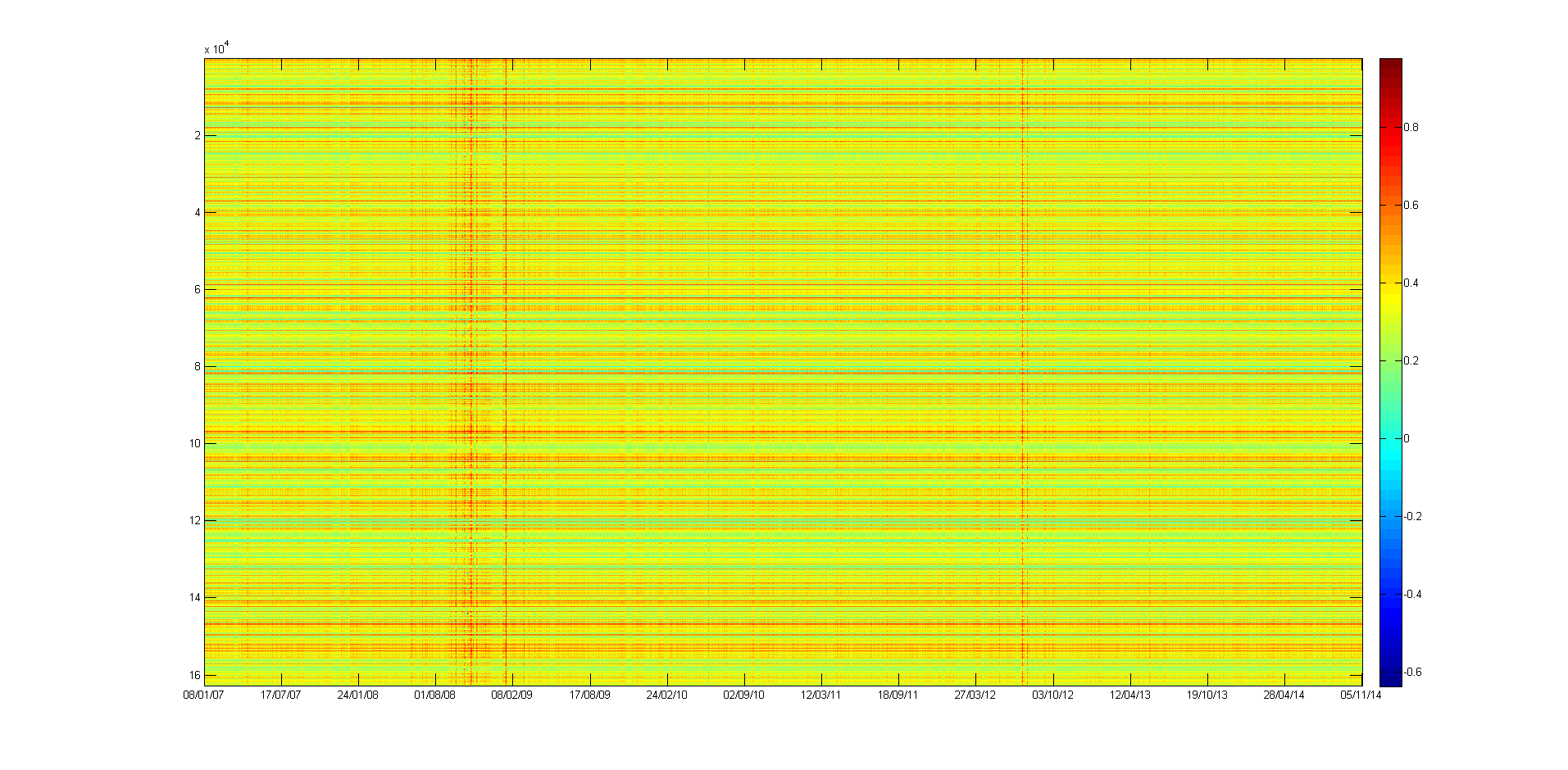}}
\caption{Posterior mean correlations of $571$ stocks of {\em STOXX Europe 600 index}}
\label{correlations}
\end{figure}

\begin{figure}[!htb]
\centering
{\includegraphics[width=\textwidth]
{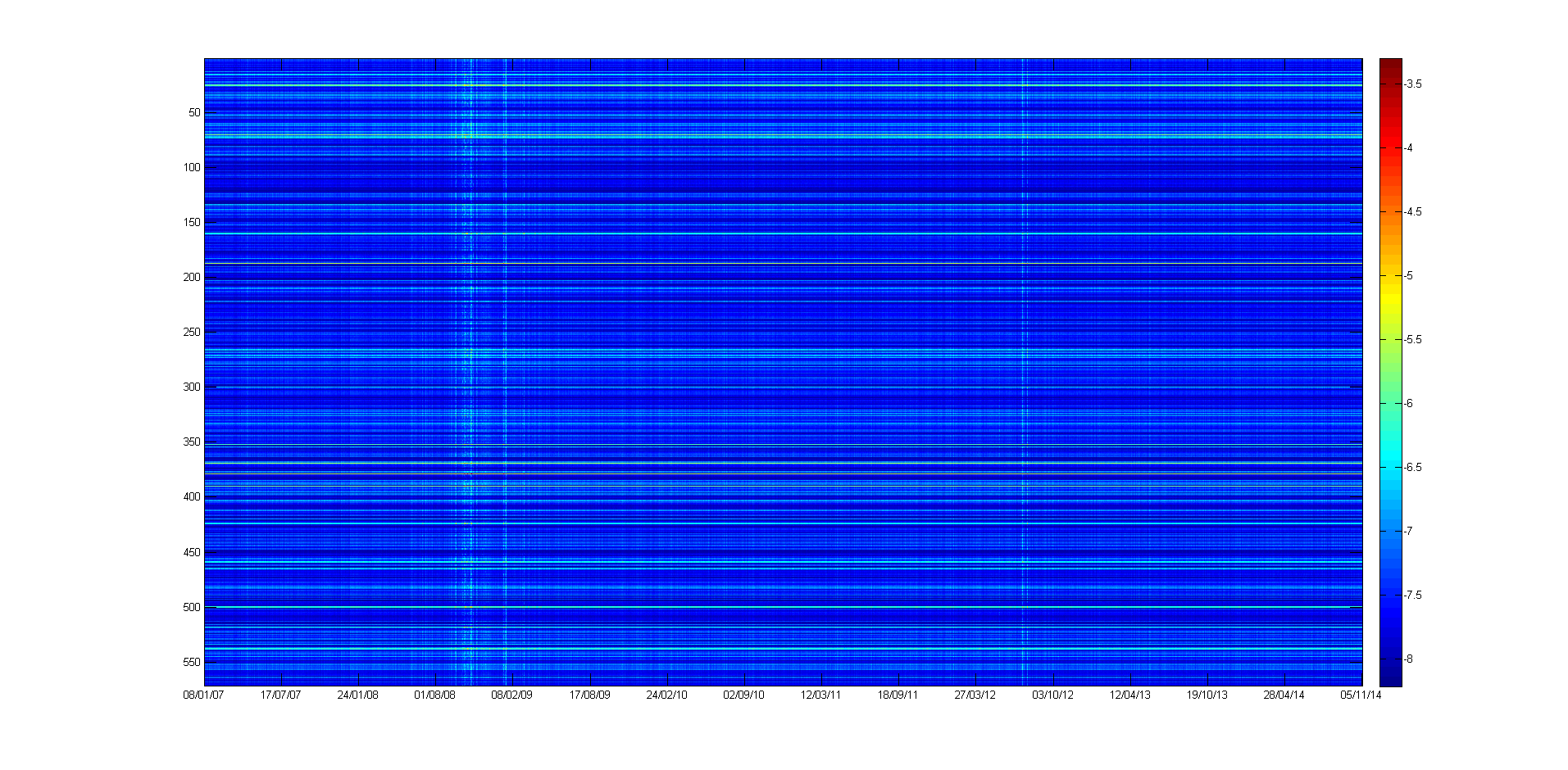}}
\caption{Posterior mean volatilities of $571$ stocks of {\em STOXX Europe 600 index}}
\label{variances}
\end{figure}

Figures \ref{correlations} and \ref{variances} are image plots of all estimated daily pairwise $571 \times 570 /2$ correlations and $571$ variances of all stocks across the whole period under study.  It is interesting that these graphs allow visual inspection of European financial contagion events by inspecting, vertically, simultaneous correlation and volatility increases.  Indeed, it is clear that our model has identified the early $2009$ financial crisis with events such as plummeting of UK banking shares, all-time high number of UK bankruptcies and eight U.S. bank failures.  Moreover, one can see the mid-$2012$ crisis after a scandal in which Barclays bank tried to manipulate the Libor and Euribor interest rates systems.

\section{Discussion}

The literature in financial econometrics suggests that univariate stochastic volatility models could be enriched by including generalisations such as allowing for non-Gaussian fat-tailed error distributions and/or jumps for the returns and leverage effects expressed through asymmetries in the relation between past negative and positive returns and future volatilities; the review papers by \citet{Asai06} and \citet{chib2009multivariate} discuss how these van be incorporated in factor models in which the factors are modelled as independent stochastic volatility processes.  We have not discussed these issues here because these extensions are not simple, especially if scalability of the MCMC algorithm is of primary concern.

We have proposed a new model and a scalable inference procedure.  If the number of assets is small, say $N=10$, one can adopt other quick inference methods such as nested Laplace approximations, see, \citet{rue2009approximate}. This is the methodology suggested and incorporated in  \citet{plataniotis2011high}, where extensive comparisons with many observation driven multivariate models is performed.  In these experiments there has been evidence that our multivariate MSV model performs better than a series of GARCH-type models.  We have not performed such experiments here mainly because estimation in multivariate GARCH models is problematic when $N$ is large and there are missing values in the returns.

\section*{Acknowledgement}
This work has been supported by the European Union, Seventh Framework Programme FP7/2007-2013 under grant agreement SYRTO-SSH-2012-320270.

\section*{Supplementary material}

The Supplementary material provides full details about the prior distributions  over the parameters 
$(\theta_h, \theta_\delta)$ and $(B,\sigma^2)$, a description of the steps for sampling these parameters, and a peudo-code for the recursive algorithm for computing the partial derivatives with respect to the rotation angles.


\bibliography{msv10}
\end{document}